\documentclass[tikz]{article}

\usepackage{PRIMEarxiv}

\usepackage[utf8]{inputenc} 
\usepackage[T1]{fontenc}    
\usepackage{hyperref}       
\usepackage{url}            
\usepackage{booktabs}       
\usepackage{amsfonts}       
\usepackage{nicefrac}       
\usepackage{microtype}      
\usepackage{lipsum}
\usepackage{fancyhdr}       
\usepackage{graphicx}       
\usepackage{csquotes}
\usepackage{subfigure}
\usepackage{mathtools} 
\usepackage{booktabs} 
\usepackage{algorithm}
\usepackage[noend]{algorithmic}
\usepackage{multirow}
\usepackage{caption}
\usepackage{subcaption}
\usepackage{amssymb}
\usepackage{graphicx}
\usepackage{xfrac}
\usepackage{xcolor}
\usepackage{ulem}
\usepackage{microtype}
\usepackage{hyperref}
\usepackage[acronym]{glossaries}
\newacronym{llm}{LLM}{Large Language Model}
\newacronym{nlp}{NLP}{Natural Language Processing}
\newacronym{emo}{EMO}{Multi-objective Evolutionary Optimization}
\newacronym{evo}{EVO}{Single-Objective Evolutionary Optimization}
\newacronym{ea}{EA}{Evolutionary Algorithm}
\newacronym{mop}{MOP}{Multi-objective Optimization Problem}
\newacronym{nsgaii}{NSGA-II}{Nondominated Sorting Genetic Algorithm II}
\newacronym{imdb}{IMDb}{Internet Movie Database}
\newacronym{api}{API}{Application Programming Interface}
\newacronym{cot}{CoT}{Chain-of-Thought}

\title{
    MOPrompt: Multi-objective Semantic Evolution for Prompt Optimization
}

\author{
Sara Câmara$^3$, Eduardo Luz$^1$, Valéria Carvalho$^1$, Ivan Meneghini$^2$ and Gladston Moreira$^1$ \\
 $^1$Computing Department, Universidade Federal de Ouro Preto, Ouro
Preto, 35402-136, Minas Gerais, Brazil.\\
$^2$Federal Institute of Minas Gerais, Ibirité-MG, Brazil.\\
$^3$Postgraduate Program in Computer Science, Federal University of Ouro Preto, Brazil.\\
    \texttt{gladston@ufop.edu.br}
}

\begin{document}
\maketitle

\begin{abstract}
    Prompt engineering is crucial for unlocking the potential of Large Language Models (LLMs). Still, since manual prompt design is often complex, non-intuitive, and time-consuming, automatic prompt optimization has emerged as a research area.
    However, a significant challenge in prompt optimization is managing the inherent trade-off between task performance, such as accuracy, and context size. Most existing automated methods focus on a single objective, typically performance, thereby failing to explore the critical spectrum of efficiency and effectiveness.
    This paper introduces the MOPrompt, a novel \gls{emo} framework designed to optimize prompts for both accuracy and context size (measured in tokens) simultaneously. Our framework maps the Pareto front of prompt solutions, presenting practitioners with a set of trade-offs between context size and performance -- a crucial tool for deploying \glspl{llm} in real-world applications.
    We evaluate MOPrompt on a sentiment analysis task in Portuguese, using Gemma-2B and Sabiazinho-3 as evaluation models.
    Our findings show that MOPrompt substantially outperforms the baseline framework. For the Sabiazinho model, MOPrompt identifies a prompt that achieves the same peak accuracy (0.97) as the best baseline solution, but with a 31\% reduction in token length.
\end{abstract}


\section{Introduction}
The emergence of powerful \glspl{llm} like GPT-4 \cite{achiam2023gpt} and Gemini \cite{team2024gemma} has revolutionized the field of \gls{nlp}. The effective use of these models is highly dependent on prompt engineering, the process of designing effective instructions to guide the model's output \cite{sahoo_systematic_2024}. However, manually crafting optimal prompts is a significant bottleneck; it is often described as a \enquote{dark art} that requires extensive trial and error.

Automating prompt optimization is a promising avenue to address this challenge \cite{zhou2022large}. In \cite{guo2023connecting}, the authors presented a framework called EvoPrompt, which combines evolutionary algorithms (EAs) with large language models (LLMs) to optimize prompts, with significant performance improvements over human-designed prompts and existing automated prompt generation methods. Yet, in \cite{stil:24}, the authors proposed an
iterative prompt evolution method to optimize the model’s performance on toxic content classification in social media. The current automatic methods concentrate exclusively on maximizing task-specific performance metrics \cite{guo2023connecting}, overlooking the context size, which is measured in tokens—the fundamental units of text processed by the model. While larger context windows can improve performance, they also require more computational resources and may lead to slower processing times. 

To tackle this issue, this work introduces \textbf{MOPrompt}, an automatic prompt optimization framework that addresses a multi-objective problem by optimizing prompts to achieve maximum accuracy and minimum context size (token length), simultaneously. As in \cite{guo2023connecting,stil:24}, we utilize LLMs to act as evolutionary operators to generate new prompts, executing semantic crossover and mutation operations to develop a diverse and contextually rich population of candidate prompts.

We pose the following research questions:
\begin{itemize}
    \item \textbf{RQ1:} Can a multi-objective evolutionary approach find prompts that are both more accurate and more token-efficient than those found by a single-objective optimization?
    \item \textbf{RQ2:} What is the impact of different prompting strategies, specifically zero-shot versus few-shot, on the multi-objective optimization process?
\end{itemize}

We conduct a set of experiments on a Portuguese sentiment classification task. We compare our approach against a baseline framework, EvoPrompt \cite{guo2023connecting}, across two distinct open-source evaluation models (Gemma-2B and Sabiazinho-3) and two prompting strategies (zero-shot and few-shot). The results demonstrate rapid convergence of MOPrompt. Using a few-shot strategy with the Sabiazinho model, the MOPrompt framework identified a prompt that achieves the same peak accuracy (0.97) as the best baseline solution, while reducing the token length by 31\%.

Our main contributions are:
\begin{itemize}
    \item A novel \gls{emo} framework, MOPrompt, for automatic prompt optimization that effectively balances task accuracy and token length (context size).
    \item An empirical study on a non-English language (Portuguese) that validates our approach and provides actionable insights into the accuracy-cost trade-off in practical prompt engineering.
\end{itemize}

\section{Related Work}
Our work is situated at the intersection of automatic prompt engineering, evolutionary algorithms, and the emerging paradigm of using \glspl{llm} as optimizers.

\paragraph{Automatic Prompt Engineering}
The field has rapidly moved from manual prompt design to automated methods \cite{korzynski2023artificial, marvin2023prompt}. Early approaches focused on discrete textual prompts, often using gradient-free optimization or search algorithms to find the best instruction from a predefined set \cite{zhou2022large}. These methods, while effective, typically optimize for a single performance metric. In contrast, our work explicitly models the trade-off between performance and cost, a crucial aspect for real-world applications that is often overlooked in academic research.

\paragraph{Evolutionary Algorithms for Prompt Optimization}
\glspl{ea} has proven to be a powerful tool for prompt optimization due to its gradient-free nature and ability to explore complex search spaces. \cite{guo2023connecting} introduced EvoPrompt to optimize discrete prompts, which connects LLMs with evolutionary algorithms. Specifically, EvoPrompt utilizes LLMs to generate new candidate prompts based on evolutionary operators (semantic operators), inspired by the genetic algorithm and differential evolution.
In \cite{baumann2024evolutionary}, the authors proposed an evolutionary multi-objective (EMO) approach tailored explicitly for prompt optimization called EMO-Prompts, using sentiment analysis capabilities as a maximization problem of the score of an emotion pair.
Our work differs in a key aspect: we focus on the fundamental trade-off between accuracy and context size. Similarly, we use an \gls{llm} itself as the core engine for performing evolutionary operations, a significant departure from using traditional, heuristic-based genetic operators.

\paragraph{LLMs as Optimizers and Operators}
A recent trend involves using \glspl{llm} not just as the target of optimization but as active components within the optimization task. Works, such as Automatic Prompt Engineer (APE) \cite{zhou2022large} and Promptbreeder \cite{fernando2023promptbreeder}, have demonstrated that \glspl{llm} can generate and refine prompts for various tasks iteratively. Also in \cite{li2023spell}, the authors propose SPELL, a semantic
prompt evolution method considering a LLM as a prompt generator. 
However, it operates in a self-referential, single-objective manner. Our proposed framework integrates the idea of an LLM-driven evolution into a formal \gls{emo} context, utilizing the \gls{llm} to execute guided crossover and mutation operations, thereby explicitly navigating the accuracy-context size trade-off.

\section{Background}
To understand our method, we first introduce the foundational concepts of \glspl{llm}, evolutionary algorithms, and multi-objective optimization.

\subsection{Large Language Models and Prompting}
\glspl{llm} are deep neural networks trained on vast amounts of text data, enabling them to understand and generate human-like text \cite{sahoo_systematic_2024}. Their behavior is steered through \textit{prompts}, which are natural language instructions. Two common prompting strategies are:
\begin{itemize}
    \item \textbf{Zero-shot prompting:} The \gls{llm} is given a direct instruction to perform a task without any examples.
    \item \textbf{Few-shot prompting:} The prompt includes a few examples (demonstrations) of the task to guide the model's output more effectively.
\end{itemize}

\subsection{Evolutionary Algorithms}
\glspl{ea} is a family of population-based metaheuristic optimization algorithms inspired by biological evolution \cite{deb2002fast}. They maintain a population of candidate solutions (individuals) that evolve over generations. Each generation involves evaluating the \textit{fitness} of individuals, selecting the best ones for reproduction, and applying genetic operators like \textit{crossover} (combining two parents to create offspring) and \textit{mutation} (introducing small, low-frequency random changes) to create a new generation.

\subsection{Multiobjective Optimization}

Many real-world problems involve the simultaneous optimization of multiple, often conflicting, objectives \cite{moreira2019guiding}. A \gls{mop} can be mathematically stated as follows:
\begin{equation}
    \min_{x \in \mathcal{X}} \mathbf{F}(x) = (f_1(x), f_2(x), ..., f_m(x))
\end{equation}
where $x$ is the vector of decision variables (or a solution) belonging to the feasible solution set $\mathcal{X}$, and $\mathbf{F}(x)$ is the vector of $m$ objective functions to be minimized.

Unlike in single-objective optimization, there is typically no single solution that is best for all objectives. The goal, instead, is to find a set of solutions representing the best possible trade-offs. The Pareto dominance principle formalizes this concept. A solution $x_a$ dominates a solution $x_b$ (denoted as $x_a \prec x_b$) if and only if $f_i(x_a) \le f_i(x_b)$ for every objective $i \in \{1, ..., m\}$ and there is at least one objective $j \in \{1, ..., m\}$ for which $f_j(x_a) < f_j(x_b)$.
A solution $x^* \in \mathcal{X}$ is called Pareto-optimal if no other solution $x \in \mathcal{X}$ dominates it.
\begin{itemize}
    \item The set of all Pareto-optimal solutions is called the Pareto set ($\mathcal{X}_E$).
    \item The image of the Pareto set in the objective space, $f(\mathcal{X}_E)$, is called the Pareto front ($\mathcal{Y}_N$).
\end{itemize}
The Pareto front represents the optimal trade-offs among the conflicting objectives. Algorithms such as \gls{nsgaii} \cite{deb2002fast} are designed to find a well-distributed and convergent approximation of this front. 

\section{The MOPrompt Framework}

We now detail our MOPrompt framework, starting with the multi-objective formulation problem, the core LLM-based genetic operators, our single-objective baseline, and concluding with the complete multi-objective approach.

\subsection{Problem Formulation}
We formally define the multi-objective prompt optimization problem. Let $p$ be a prompt from the space of all possible text-based prompts $\mathcal{P}$. We aim to find a set of Pareto-optimal prompts $P^* \subset \mathcal{P}$ that solves the following bi-objective optimization problem:
\begin{equation}
    \min_{p \in \mathcal{P}} F(p) = (f_{\text{cost}}(p), f_{\text{error}}(p))
\end{equation}
where the two objective functions to be minimized are:
\begin{itemize}
    \item \textbf{Cost:} $f_{\text{cost}}(p) = f_{\text{tokens}}(p)$, the number of tokens in prompt $p$. This function measures the computational efficiency of the prompt.
    \item \textbf{Error:} $f_{\text{error}}(p) = 1 - f_{\text{acc}}(p, D, M, S)$, the classification error rate, where $f_{\text{acc}}$ is the accuracy on a dataset $D$ using an evaluator model $M$ and a prompting strategy $S$.
\end{itemize}
This formulation aligns with standard minimization problems in \gls{emo} frameworks, such as \gls{nsgaii}.

\subsection{LLM-based Genetic Operators}
Our framework employs a generator \gls{llm}, $M_G$ (GPT-4o mini \cite{achiam2023gpt}), to execute genetic operations. Diverging from traditional heuristic-based operators that manipulate text superficially, we leverage the semantic understanding of a \gls{llm}. We've defined a unified genetic function, $GA_{LLM}$, which performs a crossover between two parent prompts ($pr_a, pr_b$) and subsequently mutates the result to produce a single offspring. This entire sequence is executed via a structured requisition to  $M_G$'s public \gls{api}, utilizing the template presented in Table \ref{tab:generator_template}. This approach facilitates the creation of new prompts that are not only syntactically valid but also semantically coherent and contextually relevant to the optimization task.
\begin{table}[!ht]
\footnotesize
\centering
\caption{LLM template for genetic operations. The generator LLM ($M_G$) is instructed to act as a prompt optimizer and perform crossover and mutation.}
\label{tab:generator_template}
\begin{tabular}{@{}p{\columnwidth}@{}}
\toprule
\textbf{Template for Generator LLM (Crossover + Mutation)} \\
\midrule
\textbf{system:} \\
 Você é um otimizador de prompts para classificação de sentimentos (positivo ou negativo). Seu papel é melhorar instruções para modelos de linguagem, gerando prompts curtos, diretos e eficazes. Gere apenas o prompt, sem explicações ou comentários adicionais: \\
\addlinespace
\textbf{user\_crossover:} \\
Prompt A: "\{prompt\_a\}" \\
Prompt B: "\{prompt\_b\}" \\
Realize uma combinação dos dois prompts, como em uma operação de crossover, mantendo clareza, coerência e o propósito original. \\
\addlinespace
\textbf{user\_mutation:} \\
 Assim como uma mutação que introduz variedade, gere uma variação deste prompt mantendo seu objetivo de classificar sentimentos com precisão:  "\{prompt\}" 
 A variação pode incluir reformulação, troca de termos ou reorganização sintática.\\
\bottomrule
\end{tabular}
\end{table}


\subsection{Multi-objective Approach: MOPrompt}
Our proposed method, MOPrompt, extends this framework to a multi-objective context. The algorithm MOPrompt aims to solve the bi-objective problem formulated in the previous section, which involves minimizing both token count and classification error.

The key difference lies in the selection phase. Instead of roulette wheel selection, MOPrompt employs the \gls{nsgaii} algorithm. At each generation, the combined population of parents and offspring is sorted into non-dominated fronts. The next generation is then populated with individuals from the best fronts. To maintain diversity along the Pareto front, a crowding distance metric is used as a tie-breaker, favoring solutions in less-crowded regions of the objective space. This process allows MOPrompt to explore the trade-off between the prompt's accuracy and context size, returning a prompt's Pareto-optimal set for the user to choose from.

\section{Experimental Setup}

\subsection{Baseline Framework: EvoPrompt}
To establish a strong baseline, we implement a version of the Evo-Prompt framework \cite{guo2023connecting}. Here, the Evo-Prompt version performs the same genetic operations in Table \ref{tab:generator_template}. This algorithm focuses solely on maximizing accuracy ($f_{\text{acc}}$). It employs a standard \gls{ea} structure where selection is performed using the roulette wheel method. The probability of a prompt being selected for reproduction is proportional to its accuracy score. While this method is effective at finding high-accuracy prompts, it inherently overlooks prompt length, often resulting in verbose and costly solutions.

\subsection{Task and Dataset}
We evaluate our methods on a binary sentiment analysis task. The dataset used is ``maritaca-ai/imdb\_pt'', a Portuguese translation of the classic \gls{imdb} movie review dataset \cite{maas-EtAl:2011:ACL-HLT2011}. For computational efficiency, we conduct our experiments on a fixed, randomly sampled subset of 100 reviews, balanced with 50 positive and 50 negative examples. This same subset is used across all experimental runs to ensure fair comparison.

\subsection{Models}
Our framework utilizes two types of models:
\begin{itemize}
    \item \textbf{Generator LLM ($M_G$):} We use \textbf{GPT-4o mini} \cite{achiam2023gpt} as the engine for our evolutionary algorithm. It is responsible for generating the initial population of prompts and for executing the crossover and mutation operations as described in Table \ref{tab:generator_template}.
    \item \textbf{Evaluator LLMs ($M$):} To assess the fitness (accuracy) of the generated prompts, we use two distinct, open-source models:
    \begin{itemize}
        \item \textbf{Gemma-2B:} A lightweight model from Google based on Gemini technology \cite{team2024gemma}.
        \item \textbf{Sabiazinho-3:} An efficient model from Maritaca AI, specifically trained for Brazilian Portuguese \cite{pires_sabia_2023}.
    \end{itemize}
\end{itemize}

\subsection{Evaluation}
We investigate four primary experimental scenarios, covering both the baseline framework, Evo-Prompt, and our proposed framework, MOPrompt, each one applied to the `Gemma-2B' and `Sabiazinho-3' evaluator models. Within each scenario, we explore both `zero-shot' and `few-shot' prompting strategies. In our experiments, we set the population size to 10 individuals, and each evolutionary run lasted for 10 generations.

\subsection{Metrics}
The performance of the prompts is measured using two primary metrics:
\begin{itemize}
    \item \textbf{Accuracy:} The proportion of correctly classified sentiment labels in our 100-review test set.
    \item \textbf{Token Count:} In \glspl{llm}, the whole input, including the prompt, is processed as a sequence of tokens, which are the fundamental units of text. The maximum number of tokens that a model can process at once is known as its context size or context window \cite{pawar2024}. The cost of using commercial \glspl{llm} and the computational load of open-source models are directly proportional to the total number of tokens processed \cite{martin2024}. Therefore, minimizing the prompt's token count is a critical objective for creating efficient, cost-effective, and scalable applications. To ensure a model-agnostic and consistent measure of prompt cost, we calculate the number of tokens for each prompt using the `TreebankWordTokenizer` from the NLTK library. This provides a standardized measure of prompt verbosity.
\end{itemize}

\section{Results and Discussion}

\begin{figure}[!ht]
  \centering
  \subfigure[Sabiazinho model]{\includegraphics[width=3.2in,height=2.5in]{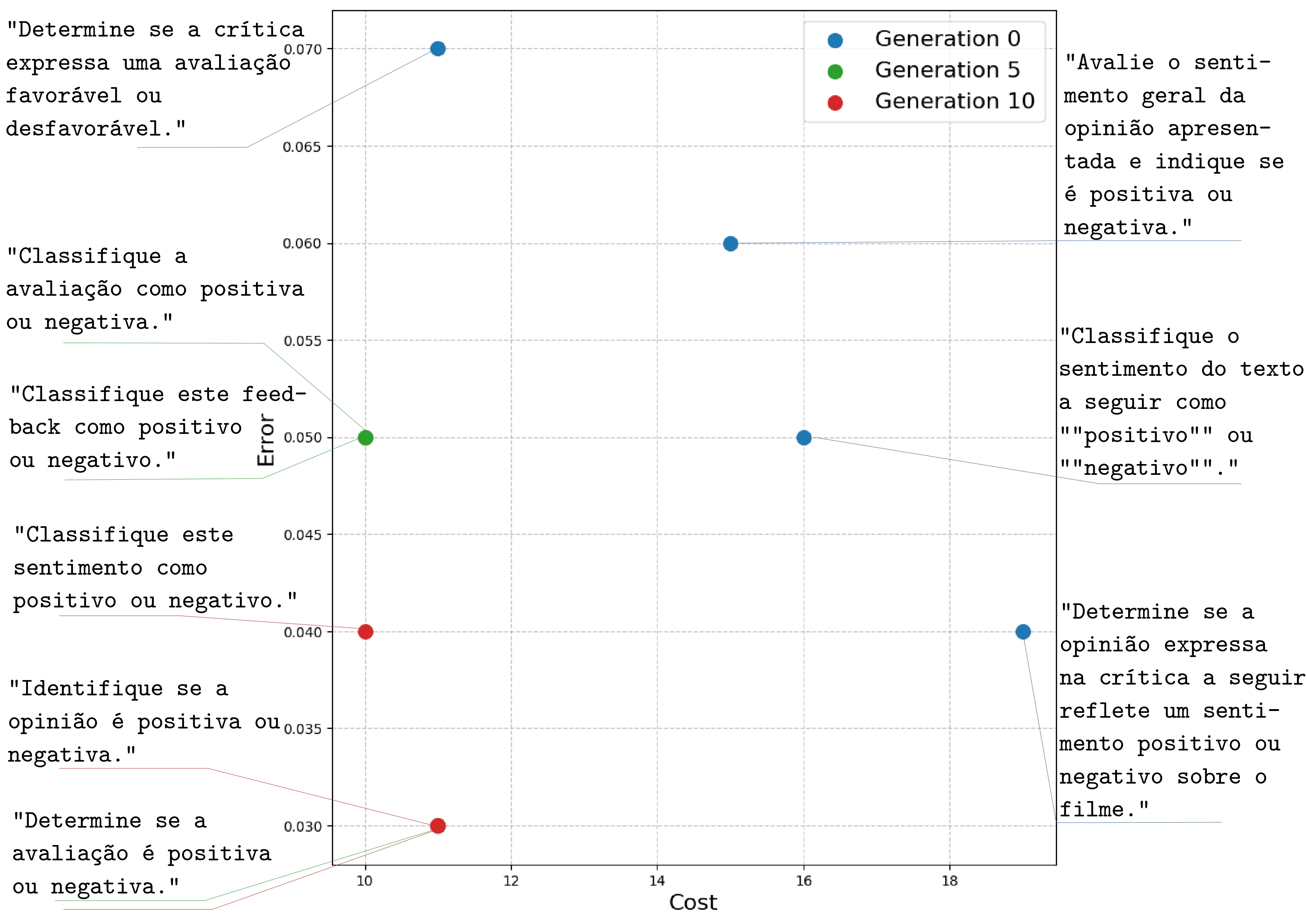}\label{fig:sabiazinho_pareto_front}}
  \subfigure[Gemma model]{\includegraphics[width=3.2in,height=2.5in]{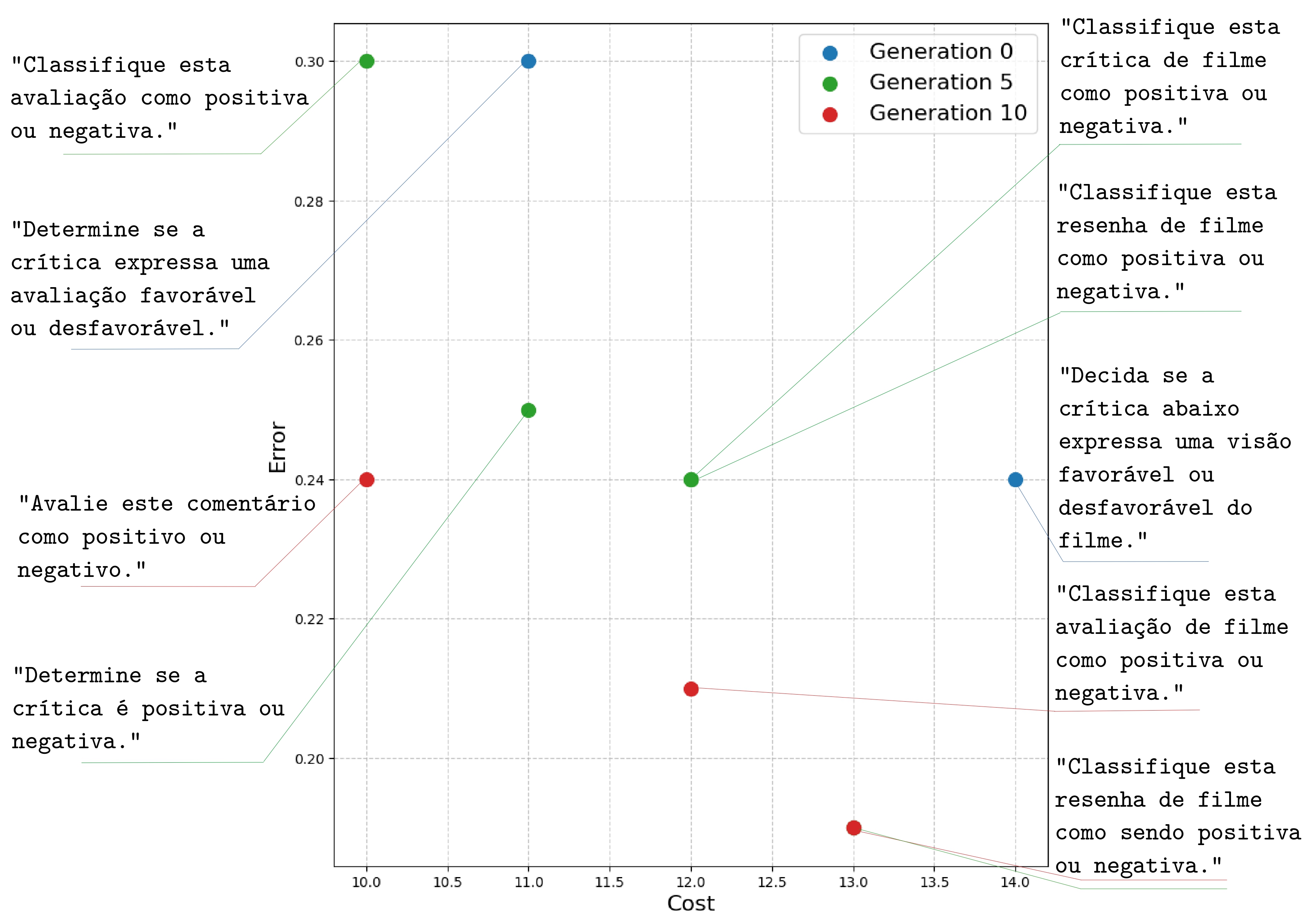}\label{fig:gemma_pareto_front}}
  \caption{Pareto Front evolution for  0, 5, and 10 generations, running the MOPrompt framework using a "few-shot" strategy.}
\end{figure}

The evolutionary dynamics of the MOPrompt framework are illustrated in Figure \ref{fig:sabiazinho_pareto_front} and Figure \ref{fig:gemma_pareto_front}, which maps the progression of the Pareto front across three key generations (0, 5, and 10) for both the Sabiazinho and Gemma models using a few-shot strategy. The visualizations confirm that the optimization process demonstrates a clear pattern of rapid convergence towards more optimal solutions.

For the Sabiazinho model, Figure \ref{fig:sabiazinho_pareto_front}, the initial prompts are scattered across a wide range of context sizes and have relatively high error rates. The evolutionary process quickly drives the population towards a more optimal state over the generations. 
A similar trend is observed with the Gemma model Figure \ref{fig:gemma_pareto_front}, although its final Pareto front reveals a more pronounced trade-off. The initial prompts in Generation 0 also start with high error rates. As the generations progress, MOPrompt successfully pushes the front toward lower error and cost. 

We now present our findings, structured around the research questions posed in the introduction. More details of the code and results are available on the repository of \href{https://anonymous.4open.science/r/mo-prompt-project-4C6B/}{Github} project.

\subsection{RQ1: Multiobjective vs. Single-Objective Performance}
Table \ref{tab:performance_summary} provides a quantitative summary of our results. The data clearly answers our first research question: the multi-objective approach, MOPrompt, consistently finds prompts that are superior to those discovered by the single-objective evoPrompt.
\begin{table*}[!ht]
\footnotesize
\centering
\caption{Performance summary comparing the MOPromot and the baseline (Evo-Prompt) framework for all scenarios.}
\label{tab:performance_summary}
\begin{tabular}{@{}ccllrr}
\toprule
\textbf{Model} & \textbf{Strategy} & \textbf{Framework} & \textbf{Highlight} & \textbf{Accuracy} & \textbf{Tokens} \\
\midrule
\multirow{6}{*}{Gemma} & \multirow{3}{*}{Few shot} & MOPrompt & Max Acc (Front) & 0.8100 & 13 \\
 & & MOPrompt & Min Tokens (Front) & 0.7600 & 10 \\
 & & Baseline & Best Accuracy & 0.8200 & 20 \\
\cmidrule{2-6}
 & \multirow{3}{*}{Zero shot} & MOPrompt & Max Acc (Front) & 0.8500 & 12 \\
 & & MOPrompt & Min Tokens (Front) & 0.7900 & 10 \\
 & & Baseline & Best Accuracy & 0.8700 & 19 \\
\midrule
\multirow{6}{*}{Sabiazinho} & \multirow{3}{*}{Few shot} & MOPrompt & Max Acc (Front) & 0.9700 & 11 \\
 & & MOPrompt & Min Tokens (Front) & 0.9600 & 10 \\
 & & Baseline & Best Accuracy & 0.9700 & 16 \\
\cmidrule{2-6}
 & \multirow{3}{*}{Zero shot} & MOPrompt & Max Acc (Front) & 0.9600 & 11 \\
 & & MOPrompt & Min Tokens (Front) & 0.9500 & 10 \\
 & & Baseline & Best Accuracy & 0.9600 & 18 \\
\bottomrule
\end{tabular}
\end{table*}

For the Gemma and Sabiazinho models, there is a minimal variation in accuracy between the MOPrompt and EVO-Prompt baseline and a significant difference in the reduction of the context size of prompts. For example, for the Gemma model in the zero-shot configuration, the best baseline prompt achieves an accuracy of 0.87 with 19 tokens. At the same time, MOPrompt finds a solution with a comparable accuracy of 0.85, but using only 12 tokens — a cost reduction of 37\%.
Another interesting observation is the difference in accuracy between the tested models. The average accuracy for both models and strategies (MOPrompt and EVO-prompt, Few-shot and Zero-shot) using the Gemma model was 83.75, while the same measurement was 96.5 for the Sabiazinho model. We believe this difference stems from the language used. All tokens were written in Brazilian Portuguese, and the Sabiazinho model is specifically designed for this language, unlike the Gemma model.

%
%
%
To understand \textit{why} MOPrompt is more effective, we qualitatively analyzed the generated prompts, as shown in Table \ref{tab:qualitative_analysis}. 
The MOPrompt framework proposed favors concise and direct instructions, while the Baseline framework (Evo-Prompt), lacking the pressure to reduce token count, often generates more verbose and less efficient prompts.
For example, the best zero-shot prompt for Gemma from Evo-Prompt is: \textit{\enquote{Classifique a opinião como positiva ou negativa, identificando se o autor demonstra aprovação ou desaprovação.}} (19 tokens). The MOPrompt that achieves a similar performance is much more direct: \textit{\enquote{Determine se a opinião apresentada é positiva ou negativa.}} (12 tokens), effectively removes redundant clauses (the "\textit{identificando...}" part) without harming performance. 
%
\begin{table}[!ht]
\centering
\caption{Qualitative analysis of the prompt solutions presented in Table \ref{tab:performance_summary}, obtained by Frameworks.}
\label{tab:qualitative_analysis} 
\resizebox{\textwidth}{!}{
\begin{tabular}{@{}llll@{}}
\toprule
\textbf{Model} & \textbf{Strategy} & \textbf{Framework}& \textbf{Prompt Highlight} \\ 
\midrule
\multirow{6}{*}{\textbf{Gemma}} & \multirow{3}{*}{Few shot} & MOPrompt & "Classifique esta resenha de filme como sendo positiva ou negativa." \\
& & MOPrompt & "Avalie este comentário como positivo ou negativo." \\
& & Baseline & "Classifique esta resenha de filme como 'positiva' (favorável) \\ 
& &  & ou 'negativa' (desfavorável)." \\
\cmidrule(l){2-4}
& \multirow{3}{*}{Zero shot} & MOPrompt & "Determine se a opinião apresentada é positiva ou negativa." \\
& & MOPrompt & "Classifique este sentimento como positivo ou negativo." \\
& & Baseline & "Classifique a opinião como positiva ou negativa, \\
& &  & identificando se o autor demonstra aprovação ou desaprovação." \\
\midrule
\multirow{6}{*}{\textbf{Sabiazinho}} & \multirow{3}{*}{Few shot} & MOPrompt & "Determine se a avaliação é positiva ou negativa." \\
& & MOPrompt & "Classifique este sentimento como positivo ou negativo." \\
& & Baseline & "Identifique se o sentimento expressado a seguir é 'positivo' ou 'negativo'." \\
\cmidrule(l){2-4}
& \multirow{3}{*}{Zero shot} & MOPrompt & "Determine se este comentário é positivo ou negativo." \\
& & MOPrompt & "Classifique a crítica como positiva ou negativa." \\
& & Baseline & "Determine se a análise a seguir reflete uma opinião positiva ou \\
& &  & negativa sobre o filme." \\
\bottomrule
\end{tabular}
 } 
\end{table}

Collectively, the evidence from both models shows that the MOPrompt methodology is promising. It actively refines and simplifies prompts over generations, validating the framework's ability to find optimal solutions that strike a balance between performance and cost.

\subsection{RQ2: Impact of Prompting Strategy}
Our second research question concerns the role of few-shot examples. Our results confirm that providing examples typically improves performance. However, the magnitude of this benefit varies significantly between models.

For the Sabiazinho model, the gap between the zero-shot and few-shot front's is modest. This indicates that Sabiazinho is an inherently strong zero-shot reasoner for this task, and the examples serve as a fine-tuning mechanism to reach peak performance. For the Gemma model, the difference is much more pronounced, suggesting a greater dependency on in-context examples to optimize the accuracy-cost trade-off. This highlights a crucial interaction between the optimization algorithm and the model's capabilities: the value of a prompting strategy is model-dependent.

\subsection{Limitations}
Despite the promising results, we acknowledge some limitations. First, while the LLM-based genetic operator is powerful, we observed that it can sometimes converge to similar prompt structures, leading to sparse Pareto fronts in some experimental runs. Future work could focus on techniques to keep greater diversity in the generated prompts. Second, our study focused on a single task (sentiment analysis) and utilized a specific set of models. Validating the generalizability of these findings across a broader range of tasks and \glspl{llm} is an essential next step.

\section{Conclusion}
In this paper, we addressed the critical challenge of balancing performance and cost in prompt engineering for \glspl{llm}. We introduced MOPrompt, a novel \gls{emo} framework that simultaneously optimizes prompts for accuracy and token efficiency. Our approach utilizes a large language model as a semantic operator to perform genetic crossover and mutation, employing Pareto dominance to evolve prompts.

Our experiments, conducted on a sentiment analysis task in Portuguese, demonstrate the clear superiority of our multi-objective approach over a strong single-objective baseline. By exploring the Pareto front of solutions, MOPrompt discovered prompts that offer significant cost reductions -- up to 31\% -- without compromising peak accuracy. Our analysis revealed that MOPrompt achieves this by generating more concise and direct instructions, effectively discovering the optimal ``language'' for each target model.

This work suggests empirical evidence for the effectiveness of \gls{emo} in prompt engineering and highlights its crucial role in optimizing the accuracy-context size trade-off. For future work, we plan to apply MOPrompt to a broader array of tasks and models, investigate methods to enhance prompt diversity, and explore its integration with more complex prompting techniques such as \gls{cot}.

\section*{Acknowledgments}
The authors would like to thank the Fundação de Amparo a Pesquisa do Estado de Minas Gerais (FAPEMIG, grant APQ-01647-22), Conselho Nacional de Desenvolvimento Científico e Tecnológico (CNPq, grants 307151/2022-0, 152613/2024-2) and Instituto Federal de Educação e Tecnologia de Minas Gerais (IFMG, grant 030/2024) for supporting the development of this study.





\bibliographystyle{main}  
\bibliography{main}

\end{document}